\documentclass[11pt]{article}

\usepackage[]{acl}

\usepackage{times}
\usepackage{latexsym}
\usepackage[T1]{fontenc}
\usepackage[utf8]{inputenc}
\usepackage{microtype}
\usepackage{inconsolata}
\usepackage{graphicx}
\usepackage{amsmath}
\usepackage{amssymb}
\usepackage{booktabs}
\usepackage{algorithm}
\usepackage{algorithmic}
\usepackage{multirow}
\usepackage{url}

\title{Is Grokking Worthwhile? Functional Analysis and Transferability of Generalization Circuits in Transformers}

\author{Kaiyu He, 
Mian Zhang, 
Peilin Wu,
Xinya Du,
Zhiyu Zoey Chen \\
Department of Computer Science\\
University of Texas at Dallas\\
\texttt{\{kaiyu.he, zhiyu.chen2\}@utdallas.edu}
}

\begin{document}
\maketitle

\begin{abstract}
While Large Language Models (LLMs) excel at factual retrieval, they often struggle with the ``curse of two-hop reasoning'' in compositional tasks. Recent research suggests that parameter-sharing transformers can bridge this gap by forming a ``Generalization Circuit'' during a prolonged ``grokking'' phase. A fundamental question arises: Is a grokked model superior to its non-grokked counterparts on downstream tasks? Furthermore, is the extensive computational cost of waiting for the grokking phase worthwhile? In this work, we conduct a mechanistic study to evaluate the Generalization Circuit's role in knowledge assimilation and transfer. We demonstrate that: (i) The inference paths established by non-grokked and grokked models for in-distribution compositional queries are identical. This suggests that the ``Generalization Circuit'' does not represent the sudden acquisition of a new reasoning paradigm. Instead, we argue that grokking is the process of integrating memorized atomic facts into an naturally established reasoning path. (ii) Achieving high accuracy on unseen cases after prolonged training and the formation of a certain reasoning path are not bound; they can occur independently under specific data regimes. (iii) Even a mature circuit exhibits limited transferability when integrating new knowledge, suggesting that ``grokked'' Transformers do not achieve a full mastery of compositional logic. Our code and data are available at: \url{https://github.com/KaiyuHe998/IsGrokkingWorthwhile}.

\end{abstract}

\section{Introduction}
Implicit reasoning refers to a model's capacity to derive new information from its stored parameters without the aid of externalized tools. A primary failure mode, often termed the ``curse of two-hop reasoning,''\citep{two_hop_curse} occurs when standard Transformers fail to compose separately learned facts. For example, a model may successfully recall that ``\textit{Barack's wife is Michelle}'' and ``\textit{Michelle was born in 1964},'' yet it remains unable to answer ``\textit{Barack's wife was born in?}'' purely through internal computation. 

\begin{figure*}[t]
    \centering
    \includegraphics[width=0.7\textwidth]{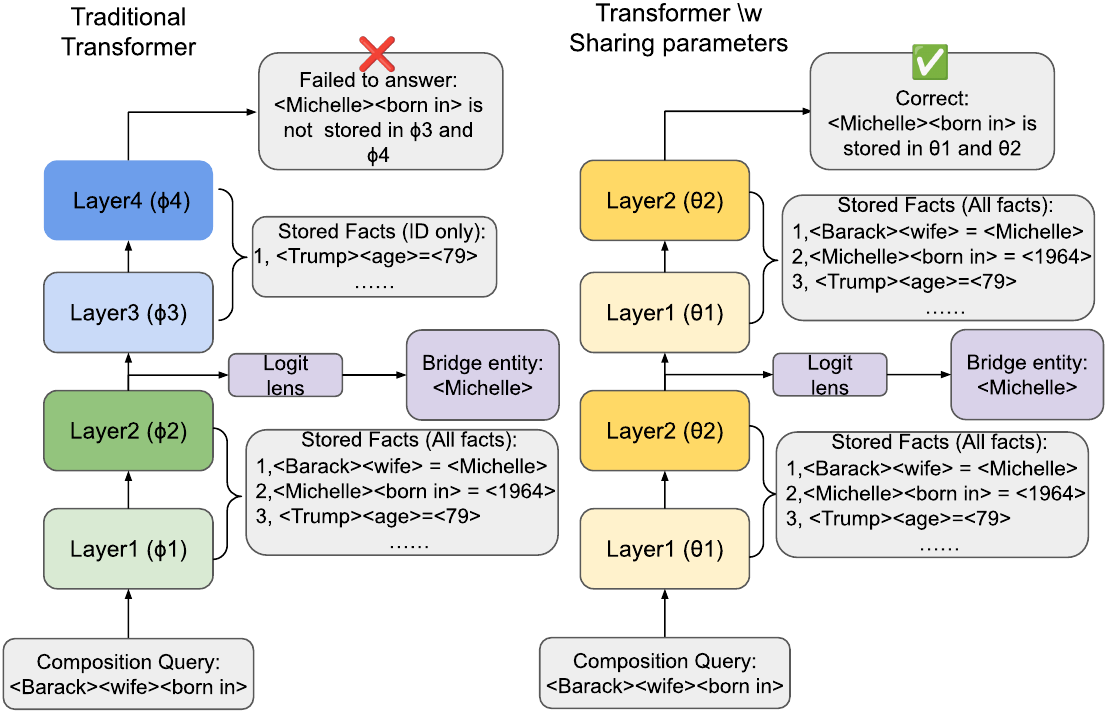}
    
    \caption{\small \textbf{Mechanistic Comparison Between traditional transformer and parameter-sharing transformer.} (Left) Standard Transformers struggle with representational mismatch across independent layers and are unable to generalize to compositional reasoning. (Right) Transformer with shared parameter resolves the bridge entity $b$ and is able to use shallow parameters to get the final answer.} 
    \label{fig:main}
    \vspace{-1.5em}
\end{figure*}

This limitation is often attributed to a \textbf{representational mismatch} in the model's internal information flow (see Figure~\ref{fig:main}). In standard Transformers, atomic facts are typically retrieved and resolved in shallow layers. When performing compositional reasoning, the bridge entity $b$ (e.g., \textit{Michelle}) emerges as an \textit{output} of these early layers. However, the subsequent deeper layers, which contain the parameters for the second hop, have not been trained to use this specific latent representation of $b$ as an \textit{input} for further retrieval \citep{Grokked_transformer}.

Existing research has demonstrated that parameter-sharing Transformers, when trained significantly beyond the point of training-label saturation, can achieve correct implicit compositional reasoning on unseen cases—a phenomenon known as \textbf{``grokking''}. Subsequent mechanistic studies have revealed that this transition is driven by the emergence of a \textbf{``Generalization Circuit''}: a specific internal computation path where the model learns to internalize the reasoning chain by explicitly recovering necessary intermediate information within its middle layers \citep{grokking_first, Grokked_transformer}. 

While prior work has sought to explain the mechanisms behind grokking \citep{Grokked_transformer, explain_grokking,nanda2023progressmeasuresgrokkingmechanistic,merrill2023a,huang2024unifiedviewgrokkingdouble,gu2025progressmeasurestheoreticalinsights,furuta2024towards,alquboj2025mechanisticinsightsgrokkingembedding,doshi2024grokgrokdisentanglinggeneralization,lv-etal-2025-language}, leveraged the phenomenon to improve model performance \citep{grokking_in_the_wild, leverage_grokking_1, DBLP:journals/corr/abs-2305-18741, furuta2024towards, africa2025learningmodularexponentiationtransformers}, and explored diverse methodologies to induce it \cite{identity_bridge, lee2024grokfastacceleratedgrokkingamplifying, xu2025let, zhou2025neuralgrokaccelerategrokkingneural, park2024accelerationgrokkinglearningarithmetic, 11272565}, a central question remains: Is a transformer trained through grokking better than its counterpart? Furthermore, does the reasoning capability acquired through grokking effectively transfer to downstream tasks or novel facts?

In this work, we leverage the state-of-the-art parameter-sharing Transformer architecture \citep{Tiny_recursive_model} on reasoning tasks to investigate the transferability of the grokked transformers on new facts. Our findings reveal that a grokked transformer with a Generalization Circuit may not be generalized as we expected:  
\vspace{-0.5em}
\paragraph{Finding 1:} Non-grokked transformer follows identical reasoning path as the grokked one on In-Distribution data, and grokking is the process of integrating memorized atomic facts into an naturally established reasoning structure. By providing sufficient atomic and compositional supervision, models can establish a stable Generalization Circuit and achieve equivalent transfer performance without undergoing the time-consuming grokking.  
\vspace{-0.5em}
\paragraph{Finding 2:} Grokking and the formation of a Generalization Circuit are not intrinsically linked, A ``grokked'' model may not necessarily reason through a Generalization Circuit, and a Generalization Circuit may not necessarily be obtained through grokking.
\vspace{-0.5em}
\paragraph{Finding 3:} Following a Generalization Circuit does not imply a full mastery of compositional reasoning, while it accelerates the assimilation of new facts in specific setups, its ability to generalize reasoning patterns across domains remains limited compared to human-like reasoning.

\begin{figure*}[t]
    \centering
    \includegraphics[width=1\textwidth]{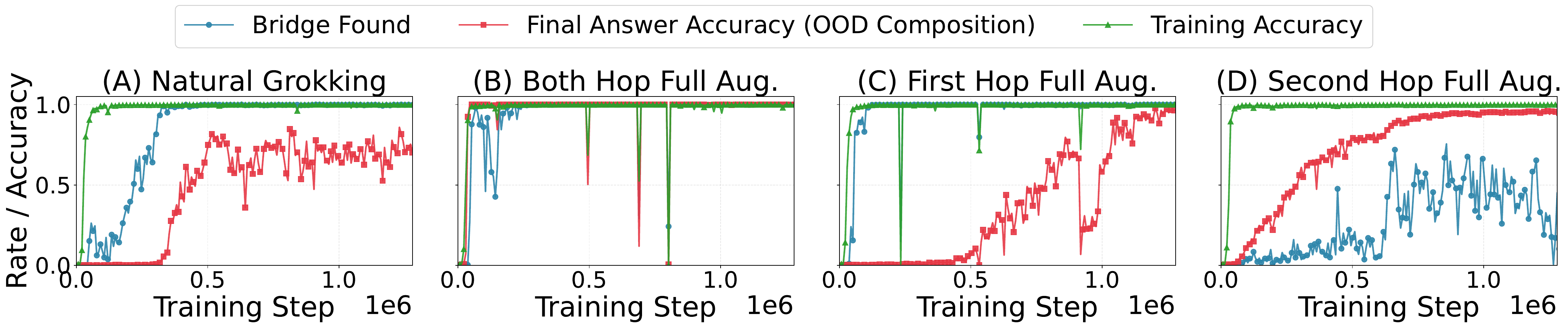}
    
    \caption{\small Transformer model ``Grokking'' under different data regime.}
    \label{fig:training curves}
    \vspace{-1.5em}
\end{figure*}

\section{Task definition \& Training data}  

We follow the compositional task definition and data format established in recent literature \citep{Grokked_transformer, ye2025how, guo2025llmsperformtwohopreasoning, brinkmann2024mechanisticanalysistransformertrained}. Models are trained on two types of queries: \textbf{atomic queries}, such as \textit{<Barack> <wife>?}'', and \textbf{compositional queries}, such as \textit{<Barack> <wife> <born in>?}''. To evaluate systematic generalization rather than surface-level memorization, we partition the set of all atomic facts into In-Distribution (ID) and Out-of-Distribution (OOD) subsets. During training, the model is exposed to all atomic queries but only those compositional queries constructed from ID facts. At test time, the model must answer compositional queries that involve OOD facts. Success in this setting indicates that the model has acquired a genuine compositional reasoning ability. Following the schema of \citet{Grokked_transformer}, we construct a synthetic dataset to ensure precise control over reasoning paths. The dataset comprises 2,000 entities and 200 unique relations, with each entity associated with 20 relations, totaling 40,000 atomic facts. We reserve 5\% (2,000 facts) as the OOD subset. More details on the construction of the dataset are shown in Appendix ~\ref{Appendix: Data Construction}.


\vspace{-0.2em}
\section{Finding 1: Grokked and non-grokked transformers follow the same reasoning path}  
\vspace{-0.2em}

\label{finding 1}

Using the logit lens \citep{logit_lens}, we visualize the maturation of Generalization Circuits in parameter-shared Transformers as they learn compositional reasoning. Specifically, we decode the hidden states across all layers with logit lens to identify the entities they represent. An inference trace is categorized as 'Bridge found' if the decoded hidden state matches the ground-truth bridge entity. We define the Generalization Circuit as the internal reasoning trace that achieves compositional reasoning by explicitly recovering the bridge entity within the model’s intermediate layers \citep{Grokked_transformer}. As shown in Figure~\ref{fig:training curves} (A), a model exhibiting natural grokking establishes three distinct stages during training on OOD compostional queries. The \textbf{first stage} is the saturation of training targets, occurring at approximately 0.1M steps. In the \textbf{second stage}, which persists for an additional 0.2M steps, the model begins to consistently recover the correct bridge entity within its intermediate layers for OOD compositional queries. The \textbf{third stage} involves the model learning to effectively utilize these intermediate representations to derive the final answer, a process that stabilizes after 1M steps. 

However, when providing compositional supervision on OOD facts, the final model trained with fewer steps follows the same reasoning path as naturally grokked models. Specifically, for \textbf{every} OOD facts appearing in the first hop of test queries, we randomly pair them with ID facts to construct \textit{OOD Facts + ID Facts} queries (\textbf{First Hop Full Augmentation}); for \textbf{every} OOD facts appearing in the second hop of test queries, we construct \textit{ID Facts + OOD Facts} queries (\textbf{Second Hop Full Augmentation}). Adding these queries ensures the model observes each OOD fact in both compositional positions, effectively equalizing the level of supervision between OOD and ID facts. Surprisingly, this supervision does not prevent the formation of the Generalization Circuit. As shown in Figure~\ref{fig:training curves}(B), within 0.3M training steps, every correct prediction on OOD compositions is achieved through successful recovery of the bridge entity in intermediate layers. Furthermore, we show in \S\ref{transferability} and Figure ~\ref{fig:Finetune plot 3_4} in Appendix ~\ref{Appendix:figure} that the model trained under full supervision exhibits identical transferability when applied to new facts compared to a natural grokked one.

These findings suggest that the ``Generalization Circuit'' does not represent the sudden acquisition of a generalized reasoning paradigm. Instead, such a reasoning path is established in the early stages of training for ID facts. Crucially, this circuit is not directly applicable to all memorized facts; rather, the subsequent grokking phase progressively integrates sparsely supervised OOD facts into this established circuit. Consequently, a model trained extensively under sparse compositional supervision and a model trained rapidly under full supervision converge on the same solution: they are \textbf{functionally equivalent} in their mechanistic approach to compositional reasoning.

\begin{figure*}[t]
    \centering
    \includegraphics[width=1\textwidth]{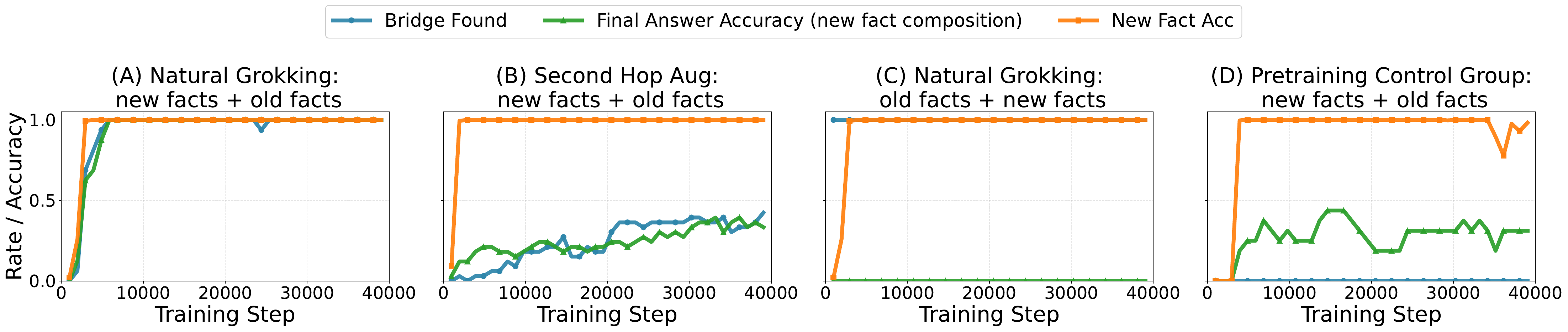}
    
    \caption{\small ``Grokked'' Transformer learning new knowledge, detailed visualization on all settings can be seen in Figure ~\ref{fig:Finetune plot 3_4} in Appendix ~\ref{Appendix:figure}}
    \label{fig:Finetune plot short}
    \vspace{-1.5em}
\end{figure*}  
\vspace{-0.2em}
\section{Finding 2: Behavioral Grokking or True Grokking?}
\vspace{-0.2em}
\label{finding 2}
First defined by \citet{grokking_first}, grokking is a phenomenon where a model suddenly achieves high accuracy on unseen cases long after saturating its performance on training targets. Previous studies have leveraged this phenomenon to improve model performance or have employed data augmentation to accelerate grokking \citep{grokking_in_the_wild}. However, we demonstrate that not all training processes that satisfy the behavioral description of grokking actually involve the formation of the intended underlying mechanism.

To investigate this, we conducted additional augmentation experiments on OOD facts. Similar to augmentation in \S~\ref{finding 1}, but this time we only did the \textbf{Second Hop Full Augmentation}. As shown in Figure~\ref{fig:training curves} (D), although the training curve fits the behavior description of grokking. Its underlying mechanism is different from Figure~\ref{fig:training curves} (A) and (B), their final answer to composition queries does not come through a correctly recovered bridge entity during the middle layer. And we will show in the \S~\ref{transferability} that such ``Fake grokked'' transformer lacks the same level of transferability compared to those transformers that really established Generalization Circuits.  

\vspace{-0.2em}
\section{Finding 3: Transferability of Generalization Circuit}  
\vspace{-0.2em}

\label{transferability}
Most evaluations of ``Grokking'' have been conducted using static datasets where the atomic facts are static. Intuitively, if a model has truly mastered a specific reasoning logic, it should exhibit human-like adaptability; once the reasoning rule is learned, it should be immediately applicable to novel facts. Despite this expectation, the downstream transferability of ``grokked'' Transformers remains largely underexplored.

\vspace{-0.2em}
\subsection{Finetuning data}
\vspace{-0.2em}

In the pre-training dataset, there are $200$ relations in total, and each entity is connected to $20$ relations. For each entity, we randomly sample relations that have not been previously used (i.e., unseen (entity, relation) pairs), and we instantiate new triples by assigning a target entity. Repeating this procedure yields entirely $2,000$ new atomic facts.

To prevent catastrophic forgetting of the original knowledge, we also retain a subset of 8,000 atomic facts from the original in-distribution (ID) training set. We then build the finetuning training split as follows:
(1) the retained $8,000$ original ID atomic facts,
(2) the $2,000$ newly added atomic facts, and
(3) all compositional queries whose supporting atomic facts are included in the retained $8,000$ ID subset (so that these compositional queries remain answerable from the included old facts).

\vspace{-0.2em}
\subsection{Limited transferability of Generalization Circuit}
\vspace{-0.2em}

We use the pre-trained checkpoints obtained in \S~\ref{finding 1} and \S~\ref{finding 2} as base models for further fine-tuning on finetuning data. This setup enables us to assess the models' capacity to assimilate new atomic facts and their efficiency in resolving associated downstream compositional queries. The resulting learning curves are illustrated in Figure~\ref{fig:Finetune plot short}.  We find that the Generalization Circuit learned during pretraining provides limited benefits. Models that have developed the intended Circuit can often recover the correct bridge entity for compositional queries involving new facts; it then leverages patterns acquired during pretraining to complete the second-hop reasoning instantly (See in Figure ~\ref{fig:Finetune plot short} (A)). 

However, when a novel fact is required in the second hop, models fail to complete the inference despite correctly recovering the bridge entity, indicating that the second component of the Generalization Circuit is less transferable. To establish a reliable circuit for new knowledge, models still require either (i) an additional grokking phase, or (ii) substantially more finetuning data covering the new facts and their compositions. Finally, ``Fake grokked'' transformers, despite achieving comparable performance on original OOD facts, fail to adapt to compositional reasoning mixing new and old facts. As shown in Figure~\ref{fig:Finetune plot short}(B), these models memorize new atomic facts but do not learn to reason over them compositionally.
\vspace{-0.3em}
\section{Conclusion}
\vspace{-0.3em}
In this work, we mechanistically re-evaluate grokking and Generalization Circuits in Transformers. Our findings challenge the view that grokking represents genuine acquisition of generalized reasoning. We show that grokked and non-grokked models follow identical reasoning paths, revealing that grokking merely integrates memorized facts into naturally established circuits. We further demonstrate a dissociation between behavioral grokking and circuit formation, and expose limited transferability to novel facts. These results suggest that the decision to pursue grokking involves a trade-off between data scarcity and computational resources: training until grokking may be necessary under sparse supervision, but with sufficient compositional data, models can stop at training saturation and achieve equivalent performance.

\section{Limitations}

Despite the mechanistic insights provided by our study, several limitations remain.

\textbf{Computational Constraints on Exploring Fake Grokking:} First, our investigation into ``fake grokking'' (behavioral grokking without circuit formation) is not exhaustive. Since the emergence of grokking phenomena typically requires extremely prolonged training cycles (often exceeding 1M steps in our settings), conducting a large-scale grid search to identify all possible augmentation strategies that might trigger circuit collapse was computationally infeasible. Consequently, while we have demonstrated that certain augmentation regimes lead to fake grokking, the precise boundary conditions and the universal principles governing this transition remain an open question. Future research is needed to systematically categorize which data distributions favor circuit formation over pattern memorization.

\textbf{Architectural Scope:} Second, our findings are specifically constrained to parameter-sharing Transformer architectures. As established in recent literature, these models are currently unique in their ability to resolve the representational mismatch that typically hinders implicit compositional reasoning. Standard Transformers with independent layer weights inherently struggle to compose separately learned facts; their deeper layers are not trained to recognize the latent representations produced by earlier layers as valid inputs for subsequent retrieval. Therefore, while our results regarding circuit formation and collapse are robust within the context of shared-parameter models, they do not necessarily extend to traditional architectures, which appear unable to achieve this form of implicit generalization even with extensive training.

\bibliography{custom}
\clearpage

\appendix  

\section{Figures}  
\label{Appendix:figure}

\begin{figure*}[h]
    \centering
    \includegraphics[width=1\textwidth]{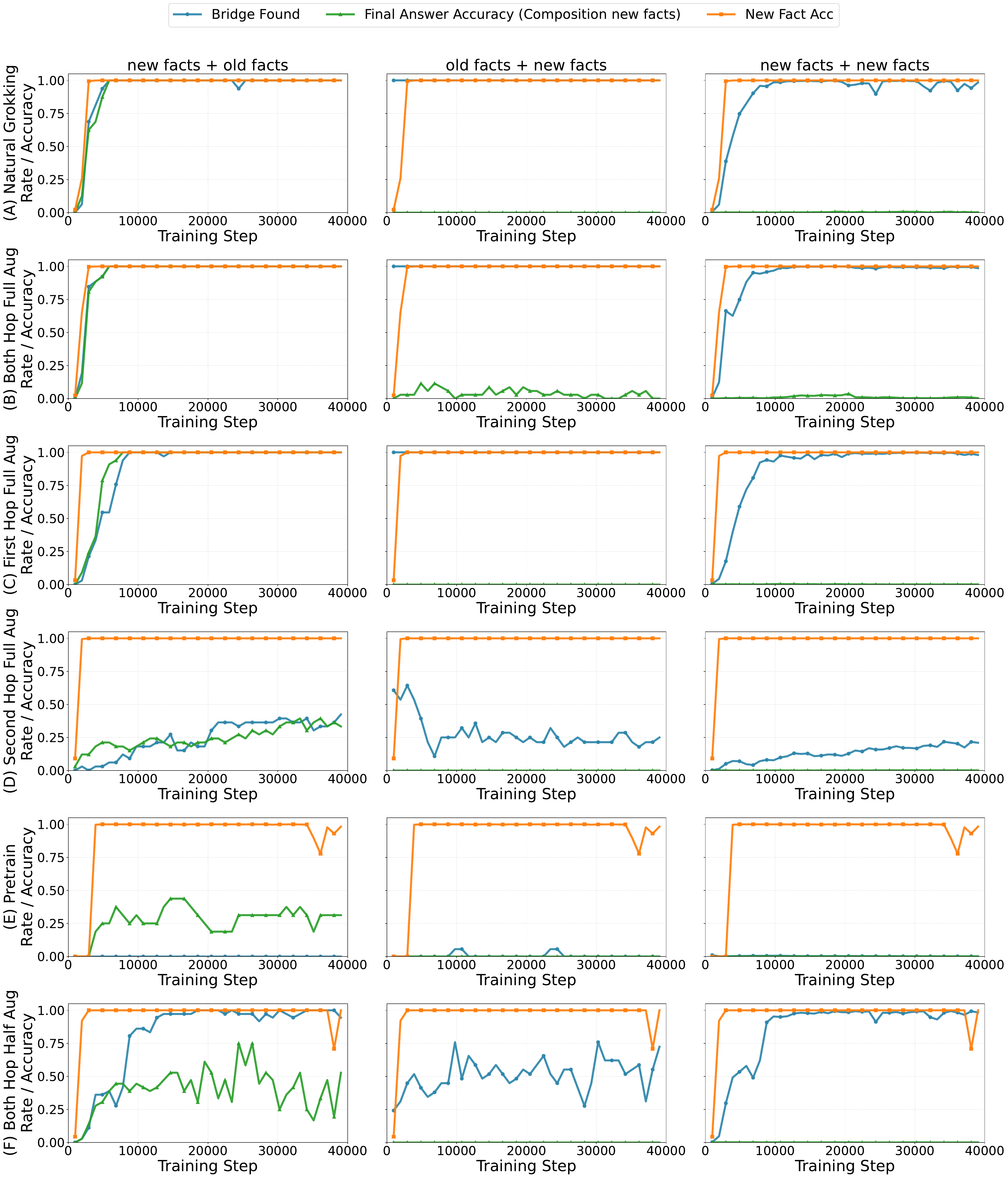}
    \caption{``Grokked'' transformer finetuned on new facts.}
    \label{fig:Finetune plot 3_4}
\end{figure*}

\clearpage

\section{Pre-training Data Construction}
\label{Appendix: Data Construction}

We adopt a synthetic data generation framework following \citet{Grokked_transformer} to ensure precise control over reasoning paths and data distributions.

\subsection{Synthetic Dataset Generation}
The dataset is modeled as a directed knowledge graph $G$, consisting of $|\mathcal{E}| = 2,000$ entities and $|\mathcal{R}| = 200$ unique relations. Each entity serves as a subject for 20 unique relations, resulting in a total of $40,000$ atomic triplets. 

To evaluate systematicity, these facts are partitioned into In-Distribution ($C_{ID}$) and Out-of-Distribution ($C_{OOD}$) subsets using a 95\%:5\% split, yielding 38,000 ID facts and 2,000 OOD facts. While the model is exposed to all atomic facts during training, compositional queries involving OOD facts are reserved strictly for testing true reasoning capability.

\subsection{Two-Hop Composition Rule}
The core task involves inducing the latent rule of two-hop composition:
\begin{equation}
    (h, r_1, b) \wedge (b, r_2, t) \Rightarrow (h, r_1, r_2, t)
\end{equation}
where $h, b, t$ denote the head, bridge, and tail entities, respectively. The model is trained to predict $t$ given $(h, r_1, r_2)$ by internally retrieving and composing the two supporting atomic facts.

\subsection{Choice of Inferred/Atomic Ratio ($\phi$)}
The ratio $\phi = |train\_inferred_{ID}| / |atomic_{ID}|$ serves as the primary determinant for the speed of grokking. We explicitly set \textbf{$\phi = 18.0$} for our experiments to facilitate efficient observation of the Generalization Circuit. 

According to the circuit efficiency hypothesis, a high ratio like $18.0$ increases the relative complexity of a memorizing circuit compared to a generalizing one. This high-density supervision incentivizes the model to transition toward a systematic reasoning circuit more rapidly. Specifically, at $\phi = 18.0$, models have been shown to achieve high generalization accuracy even before training performance saturates, providing a clear window for mechanistic analysis.

\section{Data Augmentation Detail}
\label{Appendix: Augmentation}

To evaluate the impact of supervision on the emergence of the Generalization Circuit, we implement targeted augmentation strategies that incorporate OOD facts into compositional training queries. 

\paragraph{Fact-level compositional augmentation (ID--OOD mixing).}
Given an atomic split $C_{\mathrm{ID}} \cup C_{\mathrm{OOD}}$ and an OOD compositional evaluation set $\mathcal{Q}_{\mathrm{OOD}}$ of queries $(h,r_1,r_2)\mapsto t$, we first \emph{extract the OOD hop facts that actually appear in the evaluation compositions}. For each $(h,r_1,r_2)\mapsto t \in \mathcal{Q}_{\mathrm{OOD}}$, we recover the bridge entity $b$, so that the corresponding hop facts are $(h,r_1,b)$ and $(b,r_2,t)$; we then collect the subsets $\mathcal{F}^{(1)}_{\mathrm{OOD}}=\{(h,r_1,b)\in C_{\mathrm{OOD}}\}$ and $\mathcal{F}^{(2)}_{\mathrm{OOD}}=\{(b,r_2,t)\in C_{\mathrm{OOD}}\}$ that occur as hop-1 / hop-2 components in $\mathcal{Q}_{\mathrm{OOD}}$.

\begin{itemize}
    \item \textbf{Hop-1 injection (OOD hop-1 + ID hop-2).}
    For each selected OOD hop-1 fact $(h,r_1,b)\in \mathcal{F}^{(1)}_{\mathrm{OOD}}$, we generate additional training compositions by sampling an \emph{ID} hop-2 fact $(b,r_x,t_x)\in C_{\mathrm{ID}}$ that shares the same bridge $b$, and add the inferred training example $(h,r_1,r_x)\mapsto t_x$.

    \item \textbf{Hop-2 injection (ID hop-1 + OOD hop-2).}
    For each selected OOD hop-2 fact $(b,r_2,t)\in \mathcal{F}^{(2)}_{\mathrm{OOD}}$, we generate additional training compositions by sampling an \emph{ID} hop-1 fact $(h_y,r_y,b)\in C_{\mathrm{ID}}$ that ends at the same bridge $b$, and add the inferred training example $(h_y,r_y,r_2)\mapsto t$.
\end{itemize}

\paragraph{Both hop full augmentation setting:} We expose \emph{all} OOD hop facts extracted from $\mathcal{Q}_{\mathrm{OOD}}$. Each OOD hop fact is injected into compositional training queries. Importantly, the augmentation \emph{does not} add the original OOD test compositions where \emph{both} hops are OOD; instead, every injected query is constructed to contain \emph{exactly one} OOD hop fact and one ID hop fact.

\end{document}